\documentclass[10pt,twocolumn,letterpaper]{article}

\usepackage{iccv}
\usepackage{times}
\usepackage{epsfig}
\usepackage{graphicx}
\usepackage{amsmath}
\usepackage{amssymb}

\usepackage{helvet}  
\usepackage{courier}  
\usepackage{url}  
\usepackage{soul}
\usepackage[utf8]{inputenc}
\usepackage{footnote}
\usepackage{tablefootnote}
\usepackage{comment}
\usepackage[para,online,flushleft]{threeparttable}

\usepackage{amsfonts,bm,color}
\usepackage{booktabs}       
\usepackage{amsopn}
\usepackage{empheq}
\usepackage{epstopdf}
\usepackage{float}
\usepackage{framed}
\usepackage{multirow}

\usepackage{adjustbox}
\usepackage{algorithm,algorithmicx}
\usepackage{algpseudocode}
\usepackage[small]{caption}
\algnewcommand{\LineComment}[1]{\State \(\triangleright\) #1}

\DeclareMathOperator{\sign}{sgn}

\usepackage{subcaption}

\definecolor{SL_color}{rgb}{0.858, 0.188, 0.478}

\usepackage[pagebackref=true,breaklinks=true,letterpaper=true,colorlinks,bookmarks=false]{hyperref}

\iccvfinalcopy 

\ificcvfinal\fi
\begin{document}


\title{Adversarial Robustness vs. Model Compression, or Both?}

\author{Shaokai Ye$^{1}$\thanks{Equal contribution}
\quad Kaidi Xu$^{2*}$~~~Sijia Liu$^{3}$~~~Jan-Henrik Lambrechts$^{1}$~~~Huan Zhang$^{5}$\\
Aojun Zhou$^{4}$~~~Kaisheng Ma$^{1+}$~~~Yanzhi Wang$^{2+}$~~~Xue Lin$^{2+}$\\
    $^1$IIIS, Tsinghua University \& IIISCT, China\,
    $^2$Northeastern University, USA\\
    $^3$MIT-IBM Watson AI Lab, IBM Research\\
    $^4$SenseTime Research, China \,
    $^5$University of California, Los Angeles, USA  
    }
\maketitle
\footnotetext[1]{Institute for Interdisciplinary Information Core Technology}
\let\thefootnote\relax\footnotetext{+ Corresponding Authors}
\begin{abstract}
It is well known that deep neural networks (DNNs) are vulnerable to adversarial attacks, which are implemented by adding crafted perturbations onto benign examples.
Min-max robust optimization based adversarial training can provide a notion of security against adversarial attacks.
However, adversarial robustness requires a significantly larger capacity of the network than that for the natural training with only benign examples.
This paper proposes a framework of concurrent adversarial training and weight pruning that enables model compression while still preserving the adversarial robustness and essentially tackles the dilemma of adversarial training.
Furthermore, this work studies two hypotheses about weight pruning in the conventional setting and finds that weight pruning is essential for reducing the network model size in the adversarial setting; training a small model from scratch even with inherited initialization from the large model cannot achieve neither adversarial robustness nor high standard accuracy. Code is available at \url{https://github.com/yeshaokai/Robustness-Aware-Pruning-ADMM}.



\end{abstract}

\section{Introduction}
\label{sec: intro}


Deep learning or deep neural networks (DNNs) have achieved extraordinary performance in many application domains such as image classification \cite{he2016deep,szegedy2017inception}, object detection and recognition \cite{lin2017feature,redmon2016you}, 
natural language processing \cite{collobert2008unified,manning2014stanford} and medical image analysis \cite{litjens2017survey,shi2018pairwise}.
Besides deployments on the cloud, deep learning has become ubiquitous on embedded systems such as mobile phones, IoT devices, personal healthcare wearables, autonomous driving \cite{bojarski2016end,fagnant2015preparing}, unmanned aerial systems \cite{carrio2017review,khan2018learning}, etc.

It has been well accepted that DNNs are vulnerable to adversarial attacks \cite{goodfellow2015explaining,xu2019interpreting,xu2018structured,zhao2018reinforced}, which raises concerns of DNNs in security-critical applications and may result in disastrous consequences.
For example, in autonomous driving, a stop sign may be mistaken by a DNN as a speed limit sign; malware may escape from deep learning based detection; and in authentication using face recognition, unauthorized people may escalate their access rights by fooling the DNN.

Adversarial attacks are implemented by generating adversarial examples, i.e., adding sophisticated perturbations onto benign examples, such that adversarial examples are classified by the DNN as target (wrong) labels instead of the correct labels of the benign examples. 
The adversary may have white-box accesses to the DNN where the adversary has full information about the model (e.g., structure and weight parameters) \cite{chen2018attacking,chen2018ead,zhao2018admm,xu2019topology}; or black-box accesses where the adversary can only make queries and observe outputs \cite{chen2017zoo,ilyas2018black}.  
The black-box scenarios are of particular interest in the Machine Learning as a Service (MLaaS) paradigm, specifically in some cases where DNN models trained through the cloud platform cannot be downloaded and are accessed only through the service's API. 

According to \cite{athalye2018obfuscated}, defenses that cause obfuscated gradients may provide a false sense of security and can be overcome with improved attack techniques such as backward pass differentiable approximation, expectation over transformation, and reparameterization. 
Also pointed out in~\cite{athalye2018obfuscated}, \emph{adversarial training leveraging min-max robust optimization}~\cite{madry2018towards} does not have obfuscated gradients issue and can be a promising defense mechanism.
Since that researchers have begun to notice the issue when designing new defenses,
more defenses have been proposed including adversarial training based ones ~\cite{liu2018advbnn,song2018improving,wang2018a} and others~\cite{tao2018attacks,yan2018deep,lin2019defensive}.

Min-max robust optimization based adversarial training~\cite{madry2018towards,tsipras2018robustness} can provide a notion of security against all first-order adversaries (i.e., attacks that rely on gradients of the loss function with respect to the input), by modeling an universal first-order attack through the inner maximization problem while the outer minimization still representing the training process.
However, as noted by \cite{madry2018towards}, adversarial robustness requires a significant larger architectural \emph{capacity of the network} than that for the natural training with only benign examples.
For example, we may need to quadruple a DNN model with state-of-the-art standard accuracy on MNIST for strong adversarial robustness.
In addition, increasing the network capacity may provide a better trade-off between standard accuracy of an adversarially trained model and its adversarial robustness \cite{tsipras2018robustness}.


\begin{figure*}[hbt!]
\centering
\begin{minipage}[b]{0.32\linewidth}
  \centering
  \centerline{\includegraphics[width=6.3cm]{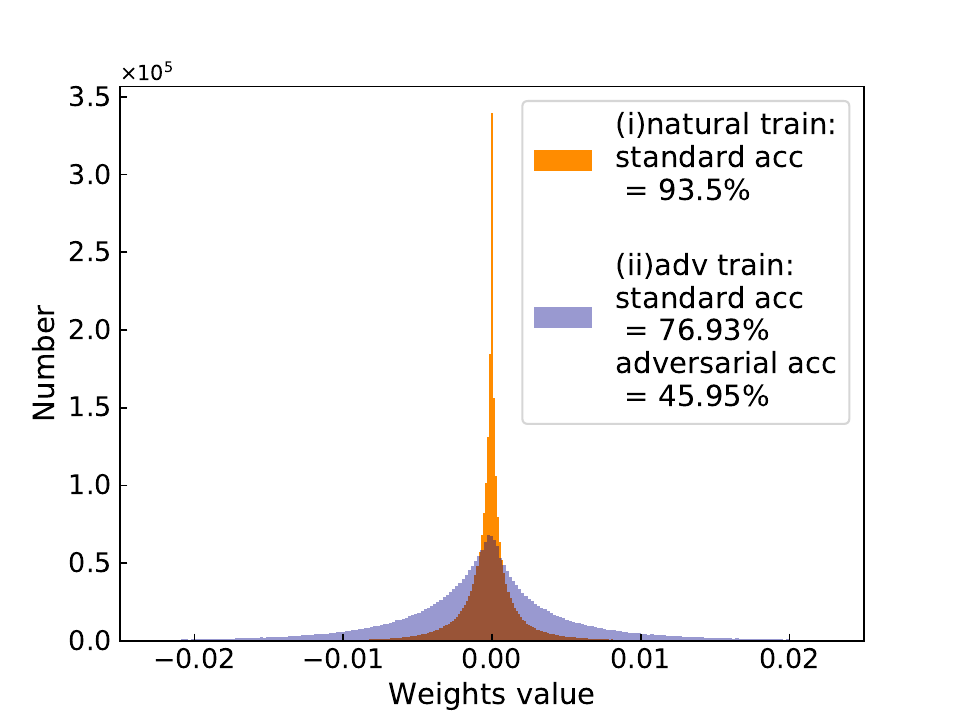}}
  \vspace{-.1cm}
  \centerline{\footnotesize(a)}\medskip
\end{minipage}
\hfill
\begin{minipage}[b]{0.32\linewidth}
  \centering
  \centerline{\includegraphics[width=6.3cm]{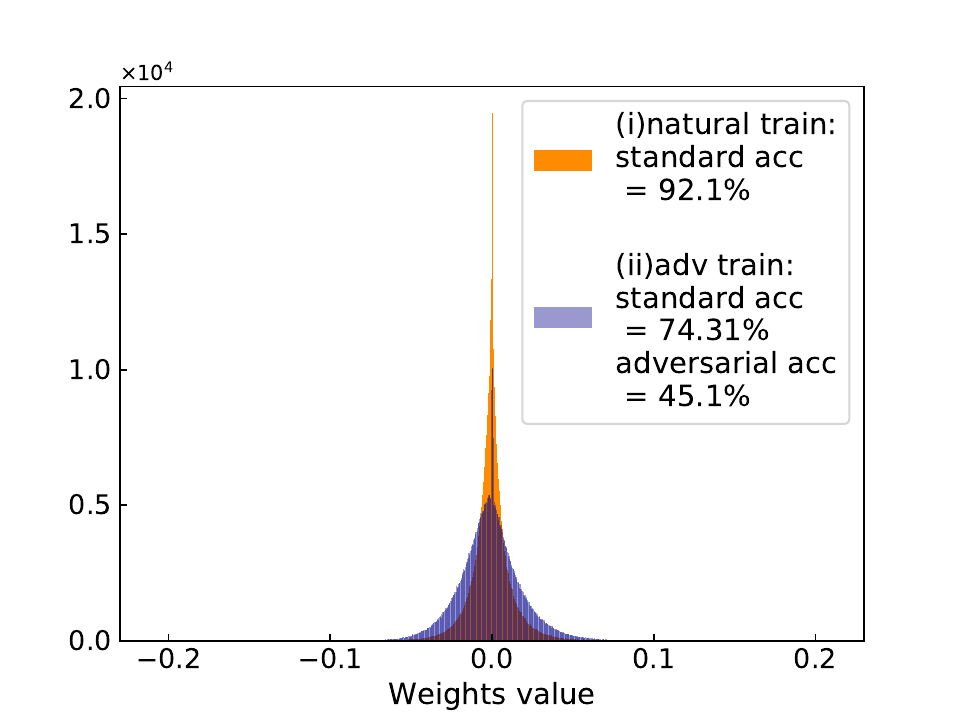}}
  \vspace{-.1cm}
  \centerline{\footnotesize(b)}\medskip
\end{minipage}
\hfill
\begin{minipage}[b]{0.32\linewidth}
  \centering
  \centerline{\includegraphics[width=6.3cm]{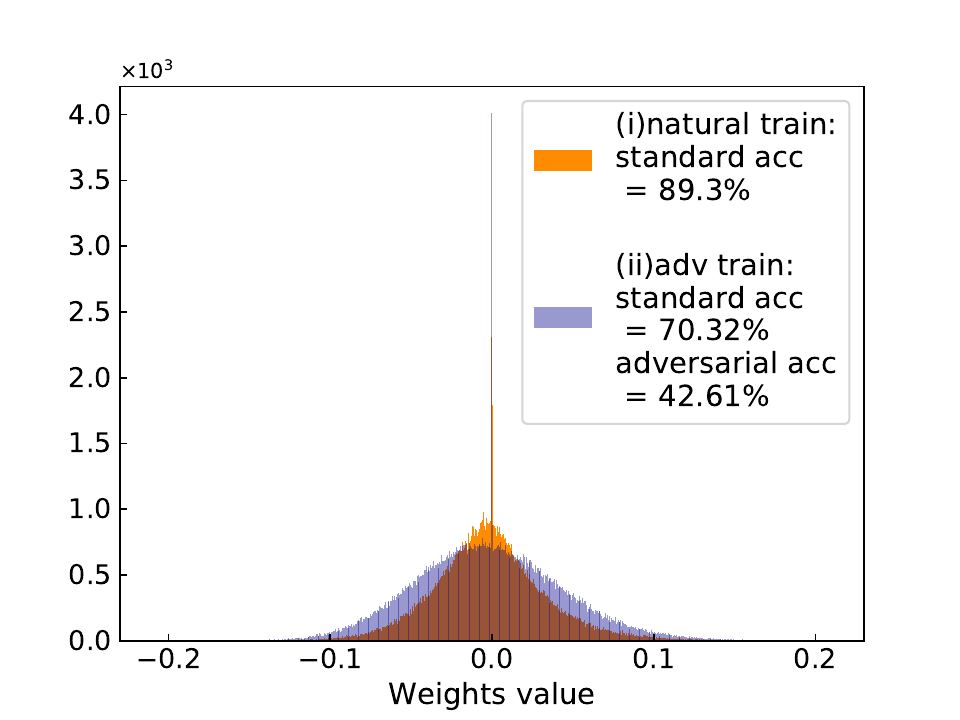}}
  \vspace{-.1cm}
  \centerline{\footnotesize(c)}\medskip
\end{minipage}
\vspace{-.1cm}
\caption{\footnotesize{Weight distribution of VGG-16 network with (a) original size, (b) 1/2 size, and (c)  1/4 size on CIFAR dataset. For each size in one subfigure, the weights are characterized for (i) a naturally trained model and (ii) an adversarially trained model. The standard accuracy and adversarial accuracy are marked with the legend.}}
\label{fig_weight_distribution}
\end{figure*}

Therefore, the required large network capacity by adversarial training may limit its use for security-critical scenarios especially in resource constrained application systems.
On the other hand, model compression techniques such as \emph{weight pruning} \textcolor{black}{\cite{han2015learning,guo2016dynamic,yang2017designing,he2017channel,wen2016learning}} have been essential for implementing DNNs on resource constrained embedded and IoT systems.
Weight pruning explores weight sparsity to prune synapses and neurons without notable accuracy degradation.
References \cite{guo2018sparse,xiao2019training} theoretically discuss the relationship between adversarial robustness and weight sparsity, but do not apply any active defense techniques in their research.
The work \cite{xiao2019training} concludes that moderate sparsity can help with adversarial robustness in that it increases the $\ell_p$ norm 
of adversarial examples (although DNNs with weight sparsity are still vulnerable under attacks).

We are motivated to investigate whether and how weight sparsity can facilitate an active defense technique i.e., the adversarial training, by relaxing the network capacity requirement.
Figure \ref{fig_weight_distribution} characterizes the weight distribution of VGG-16 network on CIFAR dataset.
We test on the original size, 1/2 size, and 1/4 size of VGG-16 network for their standard accuracy and adversarial accuracy.
We have following observations:
(i) Smaller model size (network capacity) indicates both lower standard accuracy and adversarial accuracy for adversarially trained model.
(ii) Adversarially trained model is less sparse (fewer zero weights) than naturally trained model.
Therefore, \emph{pre-pruning before adversarial training is not a feasible solution} and \emph{it seems harder to prune an adversarially trained model}.

This paper tries to answer the question of whether we can enjoy both the adversarial robustness and model compression together.
Basically, we integrate weight pruning with the adversarial training to enable security-critical applications in resource constrained systems. 

\textbf{Our Contributions:} 
We build a framework that achieves both adversarial robustness and model compression through implementing concurrent weight pruning and adversarial training.
Specifically, we use the ADMM (alternating direction method of multipliers) based pruning~\cite{zhang2018systematic,zhang2018adam} in our framework due to its compatibility with adversarial training.
More importantly, the ADMM based pruning is universal in that it supports both irregular pruning and different kinds of regular pruning, and in this way we can easily switch between different pruning schemes for fair comparison.
Eventually, our framework tackles the dilemma of adversarial training.

We also study two hypotheses about weight pruning that were proposed for the conventional model compression setting and experimentally examine their validness for the adversarial training setting.
We find that the weight pruning is essential for reducing the network model size in the adversarial setting, training a small model from scratch even with inherited initialization from the large model cannot achieve adversarial robustness and high standard accuracy at the same time.


With the proposed framework of concurrent adversarial training and weight pruning, we systematically investigate the effect of different pruning schemes on adversarial robustness and model compression. 
We find that irregular pruning scheme is the best for preserving both standard accuracy and adversarial robustness while pruning the DNN models.

\section{Related Work}
\label{sec: related}

\subsection{Adversarial Training}\label{sec:adv_train}

Adversarial training \cite{madry2018towards} uses a min-max robust optimization formulation to capture the notion of security against adversarial attacks. It does this by modeling an universal first-order attack through the inner maximization problem while the outer minimization still represents the training process.
Specifically, it solves the optimization problem:
\begin{equation}
\label{eq:minmax}
    \min_{\boldsymbol{\theta}} \quad   \mathbb{E}_{(x,y)\sim\mathcal{D}}\left[\max_{\delta\in \boldsymbol{\Delta} }L(\boldsymbol{\theta},x+\delta,y)\right]
\end{equation}
where pairs of examples $x \in \mathbb{R}^d$ and corresponding labels $y \in [k]$ follow an underlying data distribution $\mathcal{D}$; 
$\delta$ is the added adversarial perturbation that belongs to a set of allowed perturbations $\boldsymbol{\Delta}  \subseteq \mathbb{R}^d$ for each example $x$;
$\boldsymbol{\theta} \in \mathbb{R}^p$ presents the set of weight parameters to be optimized; 
and $L(\boldsymbol{\theta}, x, y)$ is the loss function, for instance, the cross-entropy loss for a DNN.

The inner maximization problem is solved by \textcolor{black}{sign-based} projected gradient descent (PGD), which presents a powerful adversary bounded by the $\ell_\infty$-ball around $x$ as:
\begin{equation}
\label{eq:pgd}
x^{t+1} = \Pi_{x+\boldsymbol{\Delta}} \left( x^t + \alpha \sign (\nabla_x L(\boldsymbol{\theta},x,y))\right)
\end{equation}
where $t$ is the iteration index,   $\alpha$ is the step size, and $\sign(\cdot)$ returns the sign of a vector.
PGD is a variant of IFGSM attack \cite{kurakin2016adversarial} and can be used with random start to add uniformly distributed noise to model $\boldsymbol{\Delta}$ during adversarial training.

\textcolor{black}{One major drawback of adversarial training is that it needs a significantly larger network capacity for achieving strong adversarial robustness than for correctly classifying benign examples only \cite{madry2018towards}.
In addition, adversarial training suffers from a more significant overfitting issue than the natural training \cite{schmidt2018adversarially}.
Later in this paper, we will demonstrate some intriguing findings related to the above mentioned observations.}

\subsection{Weight Pruning}
Weight pruning as a model compression technique has been proposed for facilitating DNN implementations on resource constrained application systems, as it explores weight sparsity to prune synapses and consequently neurons without notable accuracy degradation.
There are in general the \emph{regular pruning scheme} that can preserve the model's structure in some sense, and otherwise the \emph{irregular pruning scheme}.
Regular pruning can be further categorized as the \emph{filter pruning scheme} and the \emph{column pruning scheme}.
Filter pruning by the name prunes whole filters from a layer.
Column pruning prunes weights for all filters in a layer, \emph{at the same locations}.
Please note that some references mention channel pruning, which by the name prunes some channels completely from the filters.
But essentially channel pruning is equivalent to filter pruning, because if some filters are pruned in a layer, it makes the corresponding channels of next layer invalid \cite{he2017channel}.

In this work, we implement and investigate the filter pruning, column pruning, and irregular pruning schemes in the adversarial training setting.
Also, with each pruning scheme, we uniformly prune every layer by the same pruning ratio.
For example, when we prune the model size (network capacity) by a half, it means the size of each layer is reduced by a half.

There are existing irregular pruning work \cite{han2015learning,guo2016dynamic,yang2017designing,zhang2018systematic} and regular pruning work \cite{he2017channel,wen2016learning,zhang2018adam,lin2017runtime,liu2017learning}.
In addition, almost all the regular pruning work are actually filter pruning, except the work \cite{wen2016learning} which is the first to propose column pruning and work \cite{zhang2018adam} which can implement column pruning through an ADMM based approach.
In this work, we use the ADMM approach due to its potential for all the pruning schemes and its compatibility with adversarial training, as shall be demonstrated in the later section.

Researchers have also begun to reflect and make some hypotheses about the weight pruning.
The lottery ticket hypothesis \cite{frankle2018the} conjectures that inside the large network, a subnetwork together with their initialization makes the pruning particular effective, and together they are termed as the ``winning tickets". In this hypothesis, the original initilizaiton of the sub-network (before the large network pruning) is needed for it to achieve competitive performance when trained in isolation.
In addition, the work \cite{liu2018rethinking} concludes that training a predefined target model from scratch is no worse or even better than applying structured (regular) pruning on a large over-parameterized model to the same target model architecture.

However, these hypotheses and findings are proposed for the general weight pruning. 
In this paper, we make some intriguing observations about weight pruning in the adversarial setting, which are insufficiently explained under the existing hypotheses \cite{frankle2018the,liu2018rethinking}.

\section{Concurrent Adversarial Training and Weight Pruning}
\label{sec: framework}

In this section, we provide the framework for concurrent adversarial training and weight pruning.
We formulate the problem in a way that lends itself to the application of ADMM (alternating direction method of multipliers):
\begin{equation}
\label{admm_form}
\begin{aligned}
& \underset{ \boldsymbol{\theta}_{i}}{\text{min}}
& & \mathbb{E}_{(x,y)\sim\mathcal{D}}\left[\max_{\delta\in \boldsymbol{\Delta} }L(\boldsymbol{\theta},x+\delta,y)\right]+\sum_{i=1}^{N} g_{i}({\bf{z}}_{i}),
\\ & \textit{s.t.}
& & {\boldsymbol{\theta}_{i}}={\bf{z}}_{i}, \; i = 1, \ldots, N.
\end{aligned}
\end{equation}
Here $\boldsymbol{\theta}_{i}$ are the weight parameters in each layer.
\begin{eqnarray}g_{i}(\boldsymbol{\theta}_{i})=
\begin{cases}
 0 & \text { if } {\boldsymbol{\theta}_{i}}\in {S_i} \\ 
 +\infty & \text { otherwise}
\end{cases}
\end{eqnarray}
is an indicator function to incorporate weight sparsity constraint (different weight pruning schemes can be defined through the set $S_i$).
${\bf{z}}_{i}$ are auxiliary variables that enable the ADMM based solution.

The ADMM framework is built on the augmented Lagrangian of an equality constrained problem \cite{boyd2011distributed}. For  problem \eqref{admm_form}, the augmented Lagrangian form becomes
\begin{align}\label{eq: Lag_aug}
& \mathcal L(\{ \boldsymbol{\theta}_i \}, \{ \mathbf  z_i \}, \{ \mathbf  u_i \}) = \mathbb{E}_{(x,y)\sim\mathcal{D}}\left[\max_{\delta\in \boldsymbol{\Delta} }L(\boldsymbol{\theta},x+\delta,y)\right] \nonumber \\
& \quad  +\sum_{i=1}^{N} g_{i}({\bf{z}}_{i}) + \sum_{i=1}^N \mathbf u_i^T (\boldsymbol{\theta}_i - \mathbf z_i) + \frac{\rho}{2} \sum_{i=1}^N \|\boldsymbol{\theta}_i  - \mathbf z_i   \|_2^2,
\end{align}
where $\{ \mathbf u_i \}$ are Lagrangian multipliers associated with equality constraints of problem \eqref{admm_form}, and $\rho > 0$ is a given augmented parameter.
Through formation of the augmented Lagrangian, the ADMM framework decomposes problem (\ref{admm_form}) into two subproblems that are solved iteratively:
\begin{align}
&\{ \boldsymbol{\theta}_i^{k}\} =  \arg\min_{\{ \boldsymbol{\theta}_i\}}   \mathcal L(\{ \boldsymbol{\theta}_i \}, \{ \mathbf  z_i^{k-1} \}, \{ \mathbf  u_i^{k-1} \}), \label{eq: sub1} \\
&\{ \mathbf  z_i^{k} \} = \arg\min_{\{ \mathbf {z}_i\}}  \mathcal L(\{ \boldsymbol{\theta}_i^{k} \}, \{ \mathbf  z_i \}, \{ \mathbf  u_i^{k-1} \}) , \label{eq: sub2}
\end{align}
where $k$ is the iteration index.
The Lagrangian multipliers are updated as  ${\bf{u}}_{i}^{k}:={\bf{u}}_{i}^{k-1}+ 
\rho ( \boldsymbol{\theta}_{i}^{k}-{\bf{z}}_{i}^{k} )$.



The first subproblem \eqref{eq: sub1} is explicitly given by
\begin{equation}
\label{subproblem_1}
 \underset{ \boldsymbol{\theta}_{i} }{\text{min}}
\ \ \mathbb{E}_{(x,y)\sim\mathcal{D}}\left[\max_{\delta\in \boldsymbol{\Delta} }L(\boldsymbol{\theta},x+\delta,y)\right]+\frac{\rho}{2}  \sum_{i=1}^{N}  \| \boldsymbol{\theta}_{i}-{\bf{z}}_{i}^{k}+{\bf{u}}_{i}^{k} \|_{2}^{2}. \\
\end{equation}
The first term in the objective function of (\ref{subproblem_1}) is a min-max problem.
Same as solving the adversarial training problem in Section \ref{sec:adv_train}, here we can use the PGD adversary (\ref{eq:pgd}) with $T$ iterations and random start 
for the inner maximization problem.
The inner problem is tractable  under
an universal first-order adversary \cite{madry2018towards}.
The second convex quadratic term in (\ref{subproblem_1})  arises due to the presence of the augmented term in \eqref{eq: Lag_aug}. 
Given the adversarial perturbation $\boldsymbol{\delta}$, we can apply the stochastic gradient decent algorithm for solving the overall minimization problem.
Due to the non-convexity of the   loss function, 
the global optimality of the solution is not guaranteed. However, ADMM could offer a local optimal solution when $\rho $ is appropriately chosen since the quadratic term in \eqref{subproblem_1} is strongly convex as $\rho > 0$, which stabilizes the convergence of ADMM \cite{hong2016convergence}.

On the other hand, the second subproblem \eqref{eq: sub2} is given by
\begin{equation}\label{eqn:subtwo}
 \underset{ \{{\bf{z}}_{i} \}}{\text{minimize}}
\ \ \ \sum_{i=1}^{N} g_{i}({\bf{z}}_{i})+ \frac{\rho}{2} \sum_{i=1}^{N}  \| \boldsymbol{\theta}_{i}^{k+1}-{\bf{z}}_{i}+{\bf{u}}_{i}^{k} \|_{2}^{2}. \\
\end{equation}
Note that $g_{i}(\cdot)$ is the indicator function defined by $S_i$, thus this subproblem can be solved analytically and optimally~\cite{boyd2011distributed}. 
The optimal solution is
\begin{equation}
\label{5}
  {\bf{z}}_{i}^{k+1} = {{\bf{\Pi}}_{S_i}}(\boldsymbol{\theta}_{i}^{k+1}+{\bf{u}}_{i}^{k}),
\end{equation}
where ${{\bf{\Pi}}_{S_i}(\cdot)}$ is Euclidean projection of $\boldsymbol{\theta}_{i}^{k+1}+{\bf{u}}_{i}^{k}$ onto $S_i$. 

\begin{algorithm}
\caption{Concurrent Adversarial Training and Weight Pruning}
\begin{algorithmic}[1]
\State Input: dataset $\mathcal{D}$, ADMM iteration number $K$, PGD step size $\alpha$, PGD iteration number $T$, augmented parameter $\rho$, and sets $S_i$'s for weight sparsity constraint.
\State Output: weight parameters $\boldsymbol{\theta}$.
\For{$k =  1,2,\ldots, K$}
    \State Sample batch $(x, y)$ from $\mathcal{D}$
    \LineComment{ Solve Eq (\ref{subproblem_1}) over $\boldsymbol{\theta}$}
        \For{$t =  1,2,\ldots, T$}
            \State Solve the inner max by Eq (\ref{eq:pgd})
        \EndFor 
         \State Apply Adam optimizer on the outer min of Eq (\ref{subproblem_1}) to obtain $\{ \boldsymbol{\theta}_i^{k}\}$
    \State Solve Eq (\ref{eqn:subtwo}) using Eq (\ref{5}) to obtain $\{ \mathbf  z_i^{k} \}$
    \State ${\bf{u}}_{i}^{k}:={\bf{u}}_{i}^{k-1}+ 
\rho ( \boldsymbol{\theta}_{i}^{k}-{\bf{z}}_{i}^{k} )$.
\EndFor  
\end{algorithmic}\label{alg: framework}
\end{algorithm}

\subsection{Definitions of $S_i$ for Weight Pruning Schemes}

This subsection introduces how to use the weight sparsity constraint ${\boldsymbol{\theta}_{i}}\in {S_i}$ to implement different weight pruning scheme. For each weight pruning scheme, we first provide the exact form of ${\boldsymbol{\theta}_{i}}\in {S_i}$ constraint and then provide the explicit form of the solution (\ref{5}).
Before doing that, we reduce $\boldsymbol{\theta}_{i}$ back into the four dimensional tensor form as $\boldsymbol{\theta}_{i} \in R^{N_i \times C_i \times H_i \times W_i}$, where $N_i, C_i, H_i$, and $W_i$ are respectively the number of filters, the number of channels in a filter, the height of a filter, and the width of a filter.


\paragraph{Filter pruning}
\begin{equation}
\label{eq_filter}
{\boldsymbol{\theta}_{i}\in {{S}}_{i} := \{\boldsymbol{\theta}_{i}\mid \|\boldsymbol{\theta}_{i}\|_{n=0} \leq \alpha_i \}}.
\end{equation}
Here, $\|\boldsymbol{\theta}_{i}\|_{n=0}$ means the number of filters containing non-zero elements. To obtain the solution (\ref{5}) with such constraint, we firstly calculate $O_{n}=\| (\boldsymbol{\theta}_{i}^{k+1} +{\bf{u}}_{i}^{k})_{n,:,:,:} \|_F^2$, where $\| \cdot \|_F$ denotes the Frobenius norm. We then keep $\alpha_i$ largest values in $O_{n}$  and set the rest to zeros. 

\paragraph{Column pruning}
\begin{equation}
\label{eq_column}
{\boldsymbol{\theta}_{i}\in {{S}}_{i} := \{\boldsymbol{\theta}_{i}\mid \|\boldsymbol{\theta}_{i}\|_{c,h,w=0} \leq \beta_i \}}.
\end{equation}
Here, $\|\boldsymbol{\theta}_{i}\|_{c,h,w=0}$ means the number of elements at the same locations in all filters in the $i$th layer containing non-zero elements. To obtain the solution (\ref{5}) with such constraint, first we calculate $O_{c}=\| (\boldsymbol{\theta}_{i}^{k+1} +{\bf{u}}_{i}^{k})_{:,c,h,w} \|_F^2$. We then keep $\beta_i$ largest values in $O_{c}$  and set the rest to zeros. 

\begin{table*}[htb]
\small
 \centering
 \caption{Network structures used in our experiments. FC, M, and A mean fully connected layer, max-pooling layer, and average-pooling layer, respectively. Other numbers denote the numbers of filters in convolutional layers. We use $w$ to denote the scaling factor of a network. Each layer is equally scaled with $w$.
  } 
  \label{table_model_structure}
\begin{tabular}{c|c}
\toprule[1pt]
MNIST & 2*$w$, 4*$w$, FC(196*$w$, 64*$w$), FC(64*$w$,10) \\ \hline
CIFAR LeNet & 6*$w$, 16*$w$, FC(400*$w$, 120*$w$), FC(120*$w$, 84*$w$), FC(84*$w$,10)\\
CIFAR VGG&   4*$w$,4*$w$,M,8*$w$,8*$w$,M,16*$w$,16*$w$,16*$w$,M,32*$w$,32*$w$,32*$w$,M,32*$w$,32*$w$,32*$w$,M,A,FC(32*$w$,10)\\
CIFAR ResNet&   $b$*$w$, where $b$ denotes $1/16$ of the size of ResNet18 \cite{he2016deep}\\

\bottomrule[1pt]
\end{tabular}

\end{table*}

\begin{table*}[hbt]
 \centering
  \caption{\textbf{Natural test accuracy/adversarial test accuracy} (in $\%$) on \textbf{MNIST} of [column ii] naturally trained model with different size $w$, [column iii] adversarially trained model with different size $w$, [columns iv--vii] concurrent adversarial training and weight pruning from a large size to a small size.} 
  \label{table_mnist_pruning}
\begin{tabular}{c|c|c|cccc}
\toprule[1pt]
$w$ &nat baseline & adv baseline    & 1           & 2           & 4           & 8           \\
\midrule[1pt]
1  &98.25/0.00 & 11.35/11.35 & -           & -           & -           & -           \\
2  &98.72/0.00 & 11.35/11.35 & 11.35/11.35 & -           & -           & -           \\
4  &99.07/0.00 & 98.15/91.38 & 96.22/89.41 & 97.68/91.77 & -           & -           \\
8  &99.20/0.00 & 98.85/93.51 & \bf{97.31/92.16} & \bf{98.31/93.93} & 98.87/94.27 & -           \\
16 &99.31/0.00 & 99.02/94.65 & 96.19/87.79 & 98.07/89.95 & \bf{98.87/94.77} & \bf{99.01/95.44} \\
\bottomrule[1pt]
\end{tabular}
\end{table*}

\begin{table*}[hbt]
 \centering
  \caption{\textbf{Natural test accuracy/adversarial test accuracy} (in $\%$) on \textbf{CIFAR10 by LeNet} of [column ii] naturally trained model with different size $w$, [column iii] adversarially trained model with different size $w$, [columns iv--vii] concurrent adversarial training and weight pruning from a large size to a small size.} 
  \label{table_cifar_lenet_pruning}
\begin{tabular}{c|c|c|cccc}
\toprule[1pt]
$w$ &nat baseline &adv baseline    & 1           & 2           & 4           & 8           \\
\midrule[1pt]
1  &74.84/0.01 & 10.00/10.00 & -           & -           & -           & -           \\
2  &78.41/0.07 & 55.03/33.29 & 50.3/31.33 & -           & -           & -           \\
4  &83.36/0.19 & 65.01/36.30 & \bf{53.30/32.41} & 62.77/34.52 & -           & -           \\
8  &85.12/0.55 & 72.80/37.67 & 52.27/31.91 & 62.22/35.42 & 70.50/37.92 & -           \\
16 & 87.22/0.93 & 74.91/38.65 & 51.28/31.30 & \bf{62.10/35.55} &\bf{70.59/37.93} & \bf{71.93/39.00} \\
\bottomrule[1pt]
\end{tabular}

\end{table*}

\begin{table*}[hbt]
 \centering
  \caption{\textbf{Natural test accuracy/adversarial test accuracy} (in $\%$) on \textbf{CIFAR10 by ResNet} of [column ii] naturally trained model with different size $w$, [column iii] adversarially trained model with different size $w$, [columns iv--vii] concurrent adversarial training and weight pruning from a large size to a small size.
  } 
  \label{table_cifar_resnet_pruning}
\begin{tabular}{c|c|c|cccc}
\toprule[1pt]
$w$ &nat baseline & adv baseline    & 1           & 2           & 4           & 8           \\
\midrule[1pt]
1  & 84.23/0.00 & 57.16/34.40 & -           & -           & -           & -           \\
2  & 87.05/0.00& 71.16/42.45 & 64.53/37.90 & -           & -           & -           \\
4  & 91.93/0.00& 77.35/44.99 & 64.36/37.78 & 73.21/43.14 & -           & -           \\
8  & 93.11/0.00& 77.26/47.28 & \bf{64.52/38.01} & \bf{73.36/43.17} & 78.12/45.49 & -           \\
16 & 94.80/0.00& 82.71/49.31 & 64.17/37.99 & 71.80/42.86 & \bf{78.85/47.19} & \bf{81.83/48.00} \\
\bottomrule[1pt]
\end{tabular}
\end{table*}

\paragraph{Irregular pruning}
\begin{equation}
\label{eq_irregular}
{\boldsymbol{\theta}_{i}\in {{S}}_{i} := \{\boldsymbol{\theta}_{i}\mid \|\boldsymbol{\theta}_{i}\|_{0} \leq \gamma_i \}}.
\end{equation}
In this special case, we only constrain the number of non-zero elements in the $i$th layer filters i.e., in $\boldsymbol{\theta}_{i}$. To obtain the solution (\ref{5}), we keep $\gamma_i$ largest magnitude elements in $\boldsymbol{\theta}_{i}$ and set the rest to zeros.

Algorithm \ref{alg: framework} summarizes the framework of concurrent adversarial training and weight pruning.

In addition to ADMM based weight pruning, we also show results of post-pruning in Appendix \ref{sec: post-pruning}, with and without retraining respectively.

\section{Weight Pruning in the Adversarial Setting }\label{sec:discussion}

\begin{table}[htb]
\small
 \caption{
 \textbf{Natural test accuracy/adversarial test accuracy} (in $\%$) on \textbf{MNIST} for validating the lottery ticket hypothesis in the adversarial setting. 
  } 
  \label{table_mnist_lottery}
 \centering
\begin{tabular}{c|cccc}
\toprule[1pt]
$w$    & 1           & 2           & 4           & 8           \\
\midrule[1pt]

2   & 11.35/11.35 & -           & -           & -           \\
4   & 11.35/11.35 & 11.35/11.35 & -           & -           \\
8   & 11.35/11.35 & 97.36/90.19 & 98.64/94.66 & -           \\
16  & 11.35/11.35 & 11.35/11.35 & 98.42/91.63 & 98.96/95.49 \\
\bottomrule[1pt]
\end{tabular}
\end{table}

In this section, we \textcolor{black}{examine the performance of weight pruning in the   adversarial setting.
We obtain intriguing  results    contradictory from those  \cite{frankle2018the,liu2018rethinking} in the conventional model compression setting.
Here we specify the proposed framework of concurrent adversarial training and weight pruning by the filter pruning scheme, which is a common pruning choice to facilitate the implementation of sparse neural networks on hardware. 
Other pruning schemes will be investigated in the experiment section.
In Table \ref{table_model_structure}, we summarize all the networks tested in the paper with their  model architectures specified by the width scaling factor $w$.
}

\subsection{Weight Pruning vs Training from Scratch}\label{sec:discussion1}

An ongoing debate about pruning is whether weight pruning is actually needed and why not just training a small network from {scratch}.
To answer this question, the work~\cite{liu2018rethinking} performs a large amount of experiments to find that (i) training a large, over-parameterized model is often not necessary to obtain an efficient final model, and (ii) the meaning of weight pruning lies in searching the architecture of the final pruned model.
In another way, if we are given with a predefined target model, it makes no difference whether we reach the target model from pruning a large, over-parameterized model or we train the target model from {scratch}.
\textcolor{black}{We also remark that the above conclusions from~\cite{liu2018rethinking} are made  while performing regular pruning.
}

Although the findings in \cite{liu2018rethinking} may hold in the setting of natural training, the story becomes different in the setting of  adversarial training. 
Tables \ref{table_mnist_pruning}, \ref{table_cifar_lenet_pruning}, and \ref{table_cifar_resnet_pruning} demonstrate the \emph{natural test accuracy} / \emph{adversarial test accuracy} of natural training, adversarial training, and concurrent adversarial training and weight pruning for different datasets and networks.
Let us take Table \ref{table_mnist_pruning} as an example.
When we naturally train a network of size  $w=1$, we have 98.25\% natural test accuracy and 0\% adversarial test accuracy. 
When  we adversarially train the network of  size  $w=1$, both natural test accuracy and adversarial test accuracy become 11.35\%, which is still quite low.
It demonstrates that the network of size   $w=1$ does not have enough capacity for strong adversarial robustness.
In order to promote the adversarial robustness, we need to adversarially train the network with size of $w=4$ at least.
\textcolor{black}{Surprisingly, by leveraging  our concurrent adversarial training and weight pruning on the network of size $w=4$, we can obtain a much smaller pruned model with the target size of $w=1$ but achieve competitive natural test accuracy / adversarial test accuracy (96.22\% / 89.41\%)  compared to the adversarially trained model of size $w = 4$.}
To obtain a network of size $w=1$ with the highest natural and adversarial test accuracy, we should apply the proposed framework on the network of size $w=8$. 
Similar observations hold for Tables \ref{table_cifar_lenet_pruning} and \ref{table_cifar_resnet_pruning}.

In summary, \textcolor{black}{the value of weight pruning is essential in the adversarial training setting: it is possible to acquire a network of small model size (by weight pruning) with  both high natural test accuracy and adversarial test accuracy.
By contrast, one may lose the natural and adversarial test accuracy if the adversarial training is directly applied to  a small-size network that is not acquired from  weight pruning.
}

\subsection{Pruning to Inherit Winning Ticket or Else?}\label{sec:discussion2}


In the natural training (pruning) setting, the lottery ticket hypothesis \cite{frankle2018the} states that the meaning of weight pruning is in that the small sub-network model can inherit the initialization (the so-called ``winning ticket'') from the large model.
Or in another way, the weight pruning is meaningful only in that it provides effective initialization to the final pruned model.

\textcolor{black}{To test whether or not the lottery ticket hypothesis is valid in the adversarial setting, we perform  adversarial training under the similar experimental setup as  \cite{frankle2018the}. The natural/adversarial test accuracy results are summarized in Table\, \ref{table_mnist_lottery}, where the result in cell $w_1$-$w_2$ ($w_1 > w_2$) denotes the accuracy of an adversarially trained model of size $w_2$ using the inherited initialization from an adversarially trained model of size $w_1$. No pruning is used in Table \ref{table_mnist_lottery}.
For example, cell 4-2 in Table \ref{table_mnist_lottery} only yields 11.35\%/11.35\% accuracy.
Recall from Table\,\ref{table_mnist_pruning} that if we use our proposed framework of concurrent adversarial training and weight pruning to prune from a model with size 4 to a small model with size of 2, we can have high accuracy $97.68\% / 91.77\%$ in cell $4$-$2$ of Table\, \ref{table_mnist_pruning}.
Our results suggest that 
the lottery ticket hypothesis requires additional careful studies in the setting of adversarial training.
}


\begin{figure}[htb]
\centerline{
\begin{tabular}{cc}
      \includegraphics[width=.45\textwidth]{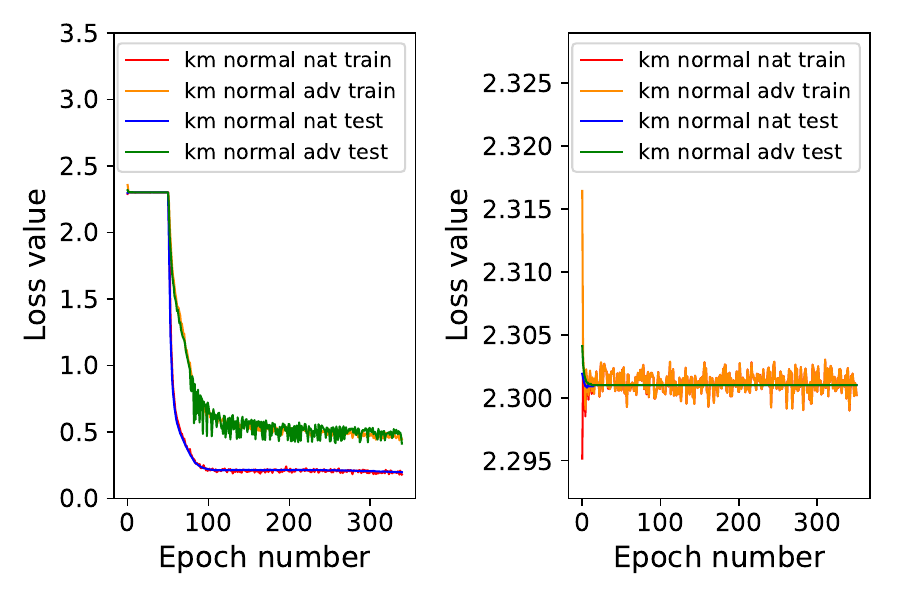}    \\
\end{tabular}}
\vspace{-0.2in}
\caption{The adversarial training loss of kaiming\_normal initialization with Adam optimizer trained from scratch on MNIST by LeNet with size of $w=1$ using different random seeds. The left subfigure is the only successful case we found and the right subfigure represents a common case.
}
  \label{fig_mnist_km_normal}
\end{figure}

Moreover, to further explore the relationship between initialization and model capacity in adversarial training, we conduct additional experiment. 
Seven different initialization methods are compared to train the smallest LeNet model ($w=1$) with 300 epochs 
using
Adam, SGD and CosineAnnealing \cite{loshchilov2016sgdr} on MNIST. 
We repeat this experiment 10 times with different random seed and report the average accuracy in Table \ref{table_mnist_init}. \textcolor{black}{As suggested by Table\,\ref{table_mnist_pruning}, adversarial training from scratch failed as $w=1,2$.
In all studied scenarios, we only find two exceptions: a) Adam with uniform initialization and b)  Adam with kaiming\_normal in which $1$ out of $10$ trials succeeds (the losses are drawn in Figure \ref{fig_mnist_km_normal}).   Even for these exceptions, the corresponding test accuracy is much worse than that of the smallest model obtained from concurrent adversarial training and weight pruning in Table\,\ref{table_mnist_pruning}. We also find that the accuracy $11.35\%$ corresponds to a saddle point that the adversarial training meets in most of cases. 
Our results in Table\,\ref{table_mnist_init} suggest that without concurrent adversarial training and weight pruning, it becomes extremely difficult to adversarially train a small model from scratch even using different initialization schemes and optimizers.
}

\subsection{Possible Benefit of  Over-Parameterization}\label{sec:discussionmore}

It is clear from Sec.\,\ref{sec:discussion1} and \ref{sec:discussion2} that in the adversarial setting, pruning from a large model is useful, which yields benefits in both natural test accuracy and adversarial robustness. By contrast, these advantages are not provided by adversarially training a small model from scratch.
Such intriguing results could be explained from the \textit{benefit of over-parameterization} \cite{zou2018stochastic,allen2018learning,allen2018convergence}, which  shows that training neural networks  possibly  reaches the global solution when the number of parameters is larger than that {is statistically} required to  fit the training data. In the similar spirit, in adversarial training setting,
the larger, over-parameterized models lead to good convergence while adversarially trained small models are stuck at the saddle points frequently. These two observations have motivated us to propose a framework that can benefit from larger models during adversarial training and at the same time reduce the models' size. As a result, the remained weights {preserve} adversarial robustness.

\section{Pruning Schemes and Transfer Attacks}
\label{sec:exp}

\begin{figure}[htb]
\centerline{
\begin{tabular}{c}
      \includegraphics[width=.48\textwidth]{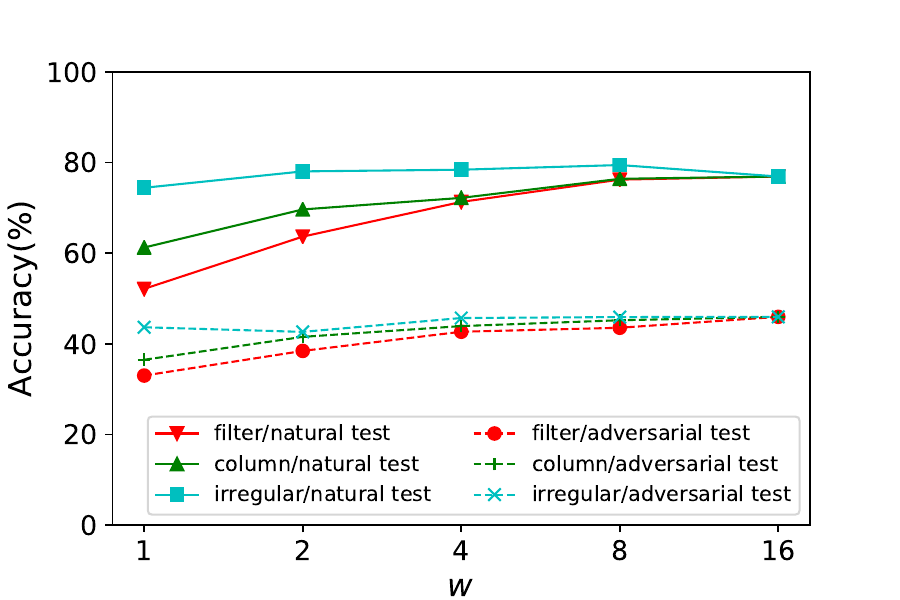}   \\
      \footnotesize (a) VGG-16\\ \vspace{-0.05in}
      \includegraphics[width=.48\textwidth]{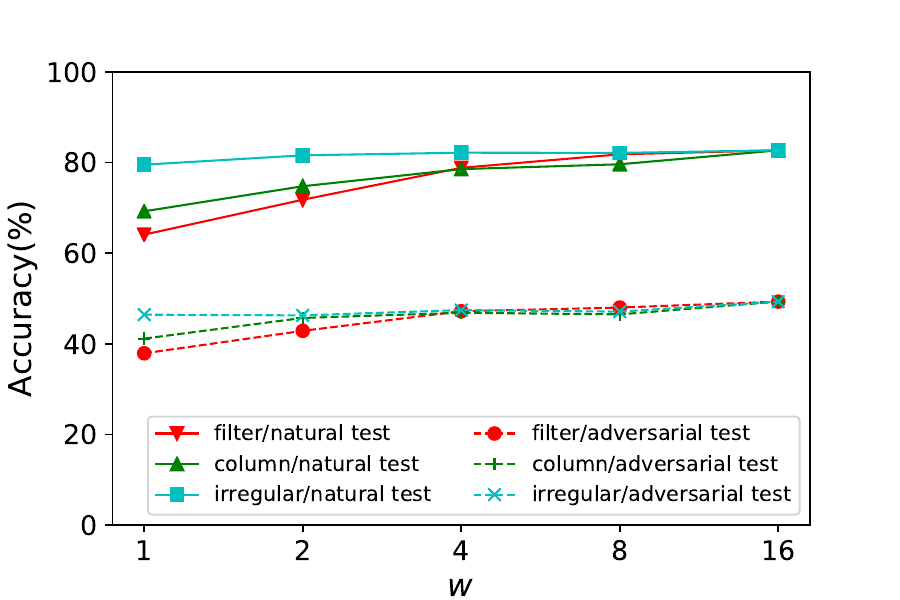}   \\
      \vspace{-0.1in}
      \footnotesize (b) ResNet-18\\
\end{tabular}}
\caption{Natural and adversarial test accuracy of the proposed framework of concurrent adversarial training and weight pruning on CIFAR10. Filter, column, and irregular pruning schemes are applied in the proposed framework respectively.
Weight pruning is performed from size of $w=16$ to sizes of $w=1,2,4,8$. The solid lines denote natural accuracy when pruning from the size of $w=16$ to different sizes, and the dashed lines denote the corresponding adversarial accuracy.
}
  \label{fig_other_pruning}
 \vspace*{-6mm}
\end{figure}

In this section, we \textcolor{black}{examine the performance of the proposed concurrent adversarial training and weight pruning under different pruning schemes (i.e., filter/column/irregular pruning) and transfer attacks.}
The proposed framework is tested on CIFAR10 using VGG-16 and ResNet-18 networks, as shown in Figure \ref{fig_other_pruning}.
As we can see, the natural and adversarial test accuracy {decrease} as the pruned size decreases.
Among different pruning schemes, the irregular pruning performs the best while the filter pruning performs the worst in both natural and adversarial test accuracy.
That is because in addition to weight sparsity, filter pruning imposes the structure constraint, which restricts the pruning granularity compared to the irregular pruning.
Moreover, irregular pruning preserves the accuracy against different pruned sizes. The reason is that the weight sparsity is beneficial to {mitigate}  the overfitting issue \cite{han2015learning}, and the adversarial training suffers a more significant overfitting than the natural training \cite{schmidt2018adversarially}.



\textcolor{black}{In Table \ref{table_mnist_cw}, we evaluate the performance of our PGD adversary based robust model against  C\&W $\ell_\infty$ attacks. {As we can see, the concurrent adversarial training and weight pruning yields the pruned model robust to transfer attacks.} In particular, the pruned model is able to achieve better adversarial test accuracy than that of the original model prior to pruning (baseline).}


Furthermore, we design a cross transfer attack experiment. Consider the baseline models in Table \ref{table_mnist_pruning}, when $w = 1,2$, the models are not well-trained so we generate adversarial examples by PGD attack from baseline model with $w = 4,8,16$ and apply them to test the pruned models. In Table \ref{table_mnist_transfer}, the results show that even the worst case in each pruned model, {the adversarial test accuracy is also higher than that of the pruned models in Table \ref{table_mnist_pruning}.}  The results imply that the model is {most vulnerable against adversarial examples generated by itself}, regardless of the size of the model.

\begin{table*}[htb]
 \centering
  \caption{Adversarial test accuracy (in $\%$) on MNIST against transfer attack from baseline models (row) when $w \in  \{4,8,16\}$ to pruned models (column) $m-n$ which means pruned from original model with $w = m$ to small model with $w = n$.
  } 
  \label{table_mnist_transfer}
\begin{tabular}{c|ccc|ccc|cc|c}
\toprule[1pt]
$w$& 16-1  & 8-1   & 4-1   & 16-2  & 8-2   & 4-2   & 16-4  & 8-4   & 16-8  \\
  \midrule[1pt]
4  & 91.70 & 93.77 & \bf{91.43} & 94.95 & 95.45 & \bf{93.55} & 97.26 & 96.56 & 97.56 \\
8  & 92.13 & \bf{93.47} & 92.06 & 94.40 & \bf{94.30} & 94.15 & 96.21 & \bf{95.27} & 96.46 \\
16 & 93.05 & 94.3  & 93.07 & \bf{94.37} & 95.72 & 94.75 & \bf{95.61} & 96.31 & \bf{95.37} \\
\bottomrule[1pt]
\end{tabular}
\end{table*}
\vspace{-0.15in}

\section{Supplementary Details of Experiment Setup}\label{sec:setup}

We use LeNet for MNIST, and LeNet, VGG-16 and ResNet-18 for CIFAR10.
The LeNet models used here follow the work \cite{madry2018towards}. Batch normalization (BN) is applied in VGG-16 and ResNet-18.
More details about the network structures are listed in Table \ref{table_model_structure}.


To solve the inner max problem in (\ref{eq:minmax}), we set PGD adversary iterations as 40 and 10, step size $\alpha$ as 0.01 and $2/255$, the $\ell_\infty$ bound as 0.3 and $8/255$ for MNIST and CIFAR respectively, and all pixel values are normalized in $[0,1]$.
We use Adam with learning rate $1\times10^{-4}$  to train our LeNet for 83 epochs as suggested by the released code of \cite{madry2018towards}. During pruning we set  $\rho = 1 \times 10^{-3}$ and $K = 30$ for Algorithm \ref{alg: framework}. Moreover, there are controversial on the baselines of CIFAR and we do the following to ensure our baselines are strong enough:
\begin{enumerate}
  \setlength{\itemsep}{0pt}

    \item We follow the suggestions by \cite{liu2018rethinking} to train our models with a larger learning rate 0.1 as initial learning rate.
    \item We train all models in CIFAR with 300 epochs and divide the learning rate by 10 times at epoch 80 and epoch 150 following the~\cite{madry2018towards}.
    \item Liu \emph{et al.}~\cite{liu2018rethinking} suggests that models trained from scratch need fair training time to compare with pruned models. Therefore, we double the training time if the loss is still descent at the end of the training.
    \item Since there is always a trade off between natural accuracy and adversarial accuracy, we report accuracy when the models achieve the lowest average loss for both natural and adversarial images on test dataset.
\end{enumerate}
Hence, we believe that in our setting, we have fair baselines for training from scratch.


\section{Conclusion}

Min-max robust optimization based adversarial training can provide a notion of security against adversarial attacks.
However, adversarial robustness requires a significant larger capacity of the network than that for the natural training with only benign examples.
This paper proposes a framework of concurrent adversarial training and weight pruning that enables model compression while still preserving the adversarial robustness and essentially tackles the dilemma of adversarial training.
Furthermore, this work studies two hypotheses about weight pruning in the conventional setting and finds that weight pruning is essential for reducing the network model size in the adversarial {setting,} and that training a small model from scratch even with inherited initialization from the large model cannot achieve adversarial robustness and high standard accuracy at the same time.
This work also systematically investigates the effect of different pruning schemes on adversarial robustness and model compression.

\section*{Acknowledgments}
This work is partly supported by the National Science Foundation CNS-1932351, Institute for Interdisciplinary Information Core Technology (IIISCT) and Zhongguancun Haihua Institute for Frontier Information Technology.

{\small
\bibliographystyle{ieeefullname}
\bibliography{egbib}
}

\clearpage

\appendix

\setcounter{section}{0}
\setcounter{figure}{0}
\makeatletter 
\renewcommand{\thefigure}{A\@arabic\c@figure}
\makeatother
\setcounter{table}{0}
\renewcommand{\thetable}{A\arabic{table}}

\section*{Appendix}

\section{Performance of Post-pruning} \label{sec: post-pruning}
We performed post-pruning for adversarially trained ResNet18 models (with variable sizes $w$) for CIFAR10 (the same setting as Table 4 in the paper). We found that without retraining, almost all cases show accuracies of 10\%/10\% (No surprise if we look at Figure 1 in the paper). Then we performed post-pruning  with retraining and show the results in following table. Consistent to {our key result}, pruning from a robust larger model gives better results than training a small model from scratch. 
{In particular, when the difference between the original size and the pruned model's size becomes large, our proposed framework  
outperforms the post-pruning-with-retraining.}
For example, the 16-to-1 case is 64.17/37.99 in Table 4 and is only 60.26/36.18 in the following table. {Furthermore, we perform additional experiments to verify the importance of ADMM.  For LeNet under FashionMNIST, post-pruning and concurrent pruning without using ADMM (proximal gradient descent is used instead) give failure cases when prune rate is large, while ADMM achieves good results under the same training  time.}
\begin{table}[hbt]
\centering
\caption{Post-pruning (with retraining) for ResNet18 on CIFAR10. Compared to ADMM method, post-pruning without retraining makes almost all models drop to 10.00/10.00.
}
\begin{adjustbox}{width=0.45\textwidth }
\begin{tabular}{c|c|c|c|c}
\toprule[1pt]
$w$  
& 1           & 2           & 4           & 8           \\
\midrule[1pt]
2  
& 64.39/38.05 & -           & -           & -           \\
4  
& 62.49/36.77 & 73.47/43.09 & -           & -           \\
8  
& 60.40/37.05 & 72.52/43.34 & 78.64/45.19 & -           \\
16 
& 60.26/36.18 & 69.47/42.14 & 78.59/46.17 & 80.79/46.4 \\

\bottomrule[1pt]
\end{tabular}
\end{adjustbox}
\label{table: post-pruning}
\end{table}
\vspace{-3mm}

\setcounter{table}{0}
\renewcommand{\thetable}{B\arabic{table}}

\section{Initialization Analysis}
The table below contains study of how initialization affects training a small robust model.
\begin{table}[htb]
\small
 \caption{\textbf{Natural test accuracy/adversarial test accuracy} (in $\%$) on \textbf{MNIST} (by LeNet with size of $w=1$)
 with seven different initialization methods and three optimizers: Adam, SGD, and CosAnneal.} 
  \label{table_mnist_init}
\begin{tabular}{c|ccc}
\toprule[1pt]
initialization   & Adam & SGD   & CosAnneal \\
                 \midrule[1pt]
uniform          & \bf{78.86/70.47}   & 11.35/11.35    & 11.35/11.35   \\
normal           & 11.35/11.35   & 11.35/11.35    & 11.35/11.35   \\
xavier\_uniform\cite{glorot2010understanding}  & 11.35/11.35   & 11.35/11.35    & 11.35/11.35   \\
xavier\_normal\cite{glorot2010understanding}   & 11.35/11.35   & 11.35/11.35    & 11.35/11.35   \\
kaiming\_uniform\cite{he2015delving} & 11.35/11.35   & 11.35/11.35    & 11.35/11.35   \\
kaiming\_normal\cite{he2015delving}   & 19.68/19.02   & 11.35/11.35    & 11.35/11.35   \\
orthogonal       & 11.35/11.35   & 11.35/11.35    & 11.35/11.35   \\
\bottomrule[1pt]
\end{tabular}
\end{table}

\setcounter{table}{0}
\renewcommand{\thetable}{C\arabic{table}}
\section{Performance against C\&W attack}
The test accuracy of our proposed framework against C\&W $\ell_\infty$ attack.
\begin{table}[htb]
 \centering
  \caption{C\&W $\ell_\infty$ adversarial test accuracy (in $\%$) by the proposed framework on MNIST by LeNet.
  } 
  \label{table_mnist_cw}
\begin{tabular}{c|c|cccc}
\toprule[1pt]
$w$  &baseline  & 1           & 2           & 4           & 8           \\
\midrule[1pt]

2   &11.35 & 11.35 & -           & -           & -           \\
4   &91.42 & 89.63 & 91.75 & -           & -           \\
8   &93.57 & 92.33 & 93.83 & 94.46  & -           \\
16  &94.78 & 89.26 & 91.34 & 95.08 & 95.62\\
\bottomrule[1pt]
\end{tabular}
\end{table}

\end{document}